\documentclass{tlp}

\usepackage{amsmath}
\usepackage{amssymb}
\usepackage{graphicx}
\usepackage{multirow}

\usepackage{microtype}
\usepackage{xspace}
\usepackage{subcaption}
\usepackage{tikz-qtree}
\usepackage{url}
\usepackage{bm}
\usepackage{times}

\newcommand{\asplarrow}{\mathrel{\mathrm{:\!\!{-}}}}
\newcommand{\weaklarrow}{\mathrel{\mathrm{:\!\!{\sim}}}}

\begin{document}

\lefttitle{Thomas Eiter, Nelson Higuera, Johannes Oetsch, and Michael Pritz}

\jnlPage{1}{15}
\jnlDoiYr{2022}
\doival{10.1017/xxxxx}

\title[A Neuro-Symbolic ASP Pipeline for Visual Question Answering]{A Neuro-Symbolic ASP Pipeline for\\ Visual Question Answering\thanks{This work was partially funded by the Bosch Center for Artificial Intelligence at Renningen, Germany.}}

\begin{authgrp}
\author{\sn{Thomas} \gn{Eiter}, \sn{Nelson} \gn{Higuera}, \sn{Johannes} \gn{Oetsch}, \sn{Michael} \gn{Pritz}}
\affiliation{%
Institute of Logic and Computation,\\ 
Vienna University of Technology \textup{(}TU Wien\textup{)}, Austria \\
\{eiter,higuera,oetsch,pritz\}@kr.tuwien.ac.at}
\end{authgrp}


\maketitle

\begin{abstract}
We present a neuro-symbolic visual question answering (VQA) pipeline for CLEVR, which is a well-known dataset that consists of pictures showing scenes with objects and questions related to them. Our pipeline covers (i) training neural networks for object classification and bounding-box prediction of the CLEVR scenes, (ii) statistical analysis on the distribution of prediction values of the neural networks to determine a threshold for high-confidence predictions, and (iii) a translation of CLEVR questions and network predictions that pass confidence thresholds into logic programs so that we can compute the answers using an ASP solver. By exploiting choice rules, we consider deterministic and non-deterministic scene encodings. Our experiments show that the non-deterministic scene encoding achieves good results even if the neural networks are trained rather poorly in comparison with the deterministic approach. This is important for building robust VQA systems if network predictions are less-than perfect. Furthermore, we show that restricting non-determinism to reasonable choices allows for more efficient implementations in comparison with  related neuro-symbolic approaches without loosing much accuracy.
This work is under consideration for acceptance in TPLP.
\end{abstract}


\section{Introduction}{\label{sec:intro}}

The goal in \emph{visual question answering} (VQA)~\citep{antol2015vqa} is to find 
the answer to a question using information from a  scene.
A system must understand the question, extract the relevant information from the corresponding scene, and perform some kind of reasoning.
\emph{Neuro-symbolic approaches} are useful in this regard as they
combine deep learning, which can be used for perception (e.g., object detection or natural language processing), with symbolic reasoning~\citep{xu2018semantic,manhaeve2018deepproblog,Yi18,yang2020neurasp,basu2020aqua,mao2019neuro}.  
As the semantics of the employed reasoning formalism is known, the way in which an answer is reached is transparent. 

We present a neuro-symbolic VQA pipeline
for the CLEVR dataset~\citep{johnson2016}
that combines \emph{deep neural networks} for perception and \emph{answer-set programming}
(ASP)~\citep{brewka2011answer} to implement the reasoning part.
The system is publicly available at
\begin{center}
\url{https://github.com/Macehil/nesy-asp-vqa-pipeline}
\end{center}

{
ASP offers a simple yet expressive modelling language and efficient solver technology.
It is in particular attractive for this task as it allows to easily express non-determinism, preferences, and defaults.  }
The scene encoding in the ASP program makes use of \emph{non-deterministic choice rules} for the objects predicted with high confidence by the network. 
This means that we do not only consider the prediction with the highest score, but also reasonable alternatives with lower ones. This allows our approach to make up for mistakes made in object classification in the reasoning component as the constraints in the program exclude choices that do not lead to an answer. 

{
For illustration, assume a scene with one red cylinder and the question  ``What shape is the red object?''.  Furthermore, assume that the neural network wrongly gives the cylinder a higher score for being blue than red. The ASP constraints enforce that an answer is derived. This entails that the right choice for the colour of the object must be red, and the correct answer ``cylinder'' is produced in the end.
}

{
While non-determinism improves the robustness of the VQA system, the downside is that, in case there are many object classes and the system is used with no restriction, it can negatively impact the reasoning performance in terms of run time. \emph{The objective of this paper is to introduce a new method for restricting non-determinism 
that is sensitive to how well networks have been trained such that efficient reasoning is facilitated.}
}

While several datasets have been published to examine the strengths and weaknesses of VQA systems~\citep{malinowski2014multi,antol2015vqa,ren2015exploring,zhu2016visual7w,johnson2016,sampat2021clevr_hyp},
{
CLEVR is an ideal test bed for the purposes of this paper, since
it is simple, well known, has detailed annotations describing the kind of reasoning each question requires, and focuses on basic object detection. 
}

We in fact 
omit natural language processing for the VQA tasks
{
because this would add a further variable and is not the focus of this work which is reasoning on top of object detection.
}
Instead, we use
\emph{functional programs} which are structured representations of the natural language questions already provided by CLEVR.

Our VQA pipeline for consists of the following stages: 
\begin{enumerate}
\item[(1)] Training neural networks for object classification and bounding-box prediction of the CLEVR scenes using the  object detection framework YOLOv3~\citep{Redmon17}.
\item[(2)] Statistical analysis on the distribution of prediction values of the neural networks to determine a threshold for high-confidence predictions defined 
as a function of mean and standard deviation of the distribution. 
\item[(3)] Translating CLEVR questions and 
network predictions that pass confidence thresholds into
logic programs so that we can compute the answers using an ASP solver.
\end{enumerate}
To determine what we regard as predictions of high confidence, we use statistical analysis on the distribution of network prediction values 
and disregard predictions that are below a threshold in Stage~(2).

Our non-deterministic scene encoding approach is inspired by the neuro-symbolic systems
NeurASP~\citep{yang2020neurasp} and DeepProbLog~\citep{manhaeve2018deepproblog}.
DeepProbLog uses ProbLog, a probabilistic extension of Prolog,  for reasoning.
There, the output of the neural network is passed to the logic program using neural-annotated disjunctions. NeurASP  uses the same bridging idea, now called neural atoms, but uses ASP instead of ProbLog. Similar to our approach, neural atoms are translated into choice rules. 
NeurASP and
DeepProbLog take multiple network predictions into account, in fact they are designed to consider all network predictions which are treated as a probability distribution. This can, as our experiments confirm, become a performance bottleneck if there are many object classes. 
Both systems feature \emph{closed-loop reasoning}, i.e., the outcome of the reasoning system can be back-propagated into the neural networks to facilitate better learning. Our pipeline is however uni-directional, as our goal is to explore the interplay between non-determinism and confidence thresholds regarding efficiency and robustness of the reasoning component.
{
We leave the learning component for future work as a scalable implementation is not  trivial in our setting and thus outside the scope of this paper.}

We compare NeurASP and the reasoning component of DeepProbLog with our approach on the CLEVR data. Indeed, limiting non-determinism of neural network outputs in ASP programs to reasonable choices leads to a drastic performance improvement in terms of run time with only little loss in accuracy and is thus important for efficient reasoning. 
Furthermore, 
our experiments show that our system performs well 
even if the neural networks are trained rather poorly and
predictions by the network are less-than perfect. This is important for robust reasoning as even well-trained networks can be negatively affected by noise or if settings like illumination change.

The remainder of this paper is organised as follows. We first 
review  ASP and CLEVR in Section~\ref{sec:bg}. 
Our VQA pipeline using ASP and confidence-thresholds is detailed in Section~\ref{sec:framework}.
Afterwards, we present an experimental evaluation of our approach in Section~\ref{sec:experiments},
discuss further relevant related work in Section~\ref{sec:rel},
and  conclude in Section~\ref{sec:concl}.

\section{Background}{\label{sec:bg}}
We next provide  preliminaries on ASP and  
background on the CLEVR dataset.

\subsection{Answer-Set Programming}

Answer-set programming (ASP) is a declarative problem solving
paradigm, where a problem is encoded as a logic program such that its
\emph{answer sets}\/ (which are special models) correspond to the
solutions of the problem and are computable using ASP solvers, e.g., from \url{potassco.org} or \url{www.dlvsystem.com}.
We just briefly recall important ASP concepts; 
and refer for more details to the literature~\citep{brewka2011answer,DBLP:series/synthesis/2012Gebser}.

An \emph{ASP program} is a finite set of \emph{rules}\/ $r$ of 
the form
{
\begin{align}\label{eq:rule}
a_1 \,| \dots |\, a_k\ \asplarrow\ b_1,
\ldots,\ b_m,\ not\ c_1, \ldots,\ \mathit{not}\ c_n, \quad k,m,n \geq 0\,
\end{align}
}
\noindent 
where all $a_i$, $b_j$, $c_l$ are first-order atoms and $\mathit{not}$
is \emph{default negation}; we denote by $H(r)=\{ a_1,\ldots, a_k \}$
and $B(r)= B^+(r) \cup \{ \mathit{not}\ c_j \mid c_j \in B^-(r)\}$
the head and body of $r$, respectively, where $B^+(r)=\{ b_1, \ldots,
b_m\}$ and $B^-(r)= \{ c_1, \ldots, c_n \}$. Intuitively, $r$ says that if all atoms in $B^+(r)$ are true and
there is no evidence that some atom in $B^-(r)$ is true, then
some atom in $H(r)$ must be true.
If $m=n=0$ and $k=1$, then $r$ is a \emph{fact} (with $\asplarrow$ omitted);
if $k=0$, $r$ is a \emph{constraint}.

An \emph{interpretation} $I$ is a set of ground (i.e., variable-free) atoms. 
It satisfies a ground rule $r$ if $H(r) \cap I \neq \emptyset$
whenever $I \subseteq B^+(r)$ and $I\cap B^-(r)=\emptyset$; $I$
is a model of a ground program $P$ if $I$ satisfies each $r\in P$, and 
$I$ is an \emph{answer-set} of $P$ if in addition no $J\subset I$ is a
model of the Gelfond-Lifschitz reduct of $P$ w.r.t.~$I$~\citep{gelfond1991classical}.  

Models and answer sets of a program $P$ with variables are defined in terms
of the grounding of $P$ (replace each rule by its possible instances over
the Herbrand universe).

We will also use \emph{choice rules} and {\em weak constraints}, which are 
of the respective forms
{
\begin{align}
\label{choice}
i\,\{a_1;\dots;a_n\}\,j \asplarrow b_1,\ldots,b_m,\ not\ c_1,\ldots,\ not\ c_n \\
\label{wc}
\weaklarrow b_1,\dots,b_m,\ \mathit{not}\ c_1, \ldots ,\ \mathit{not}\ c_n.\ [w,t]
\end{align}}
\noindent 
Informally, (\ref{choice}) says that
when the body is satisfied, at least $i$ and at most $j$ atoms from
$\{a_1,\dots,a_n\}$ must be true in an answer set $I$, while (\ref{wc}) 
contributes tuple $t$ with costs $w$, which is an integer number, to a cost function,
when the body is satisfied in
$I$, rather than to eliminate 
$I$; 
the answer set $I$ is optimal, if the total cost of all such tuples is minimal.

\subsection{The CLEVR Dataset}\label{sec:clevr}

\begin{figure}
     \centering
     \begin{subfigure}{0.3\textwidth}
         \centering
         \includegraphics[width=0.95\textwidth]{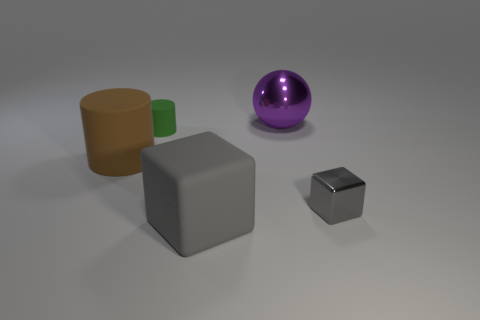}
         \caption*{\textbf{Q}: Is there a big brown object of the same shape as the green thing? \\ \textbf{A}: Yes}
     \end{subfigure}
     \hfill
     \begin{subfigure}{0.3\textwidth}
         \centering
         \includegraphics[width=0.95\textwidth]{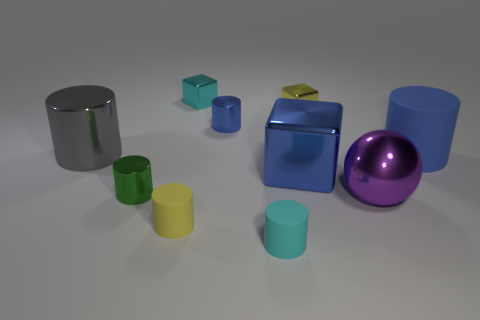}
         \caption*{\textbf{Q}: How many large things are either cyan metallic cylinders or yellow blocks? \\ \textbf{A}: 0}
     \end{subfigure}
     \hfill
     \begin{subfigure}{0.3\textwidth}
         \centering
         \includegraphics[width=0.95\textwidth]{figures/CLEVR_val_000002.png}
         \caption*{\textbf{Q}: The tiny shiny cylinder has what color? \\ \textbf{A}: Brown \\ \hphantom{...} }
     \end{subfigure}
    \caption{Three scenes and question-answer pairs from the CLEVR validation set.}
    \label{fig:three_clevr_scenes}
\end{figure}

CLEVR~\citep{johnson2016} is a dataset designed to test and diagnose the reasoning capabilities of VQA systems.\footnote{\url{https://cs.stanford.edu/people/jcjohns/clevr/}.}
It consists of pictures showing scenes with objects and questions related to them; there are about ten questions per image.
The dataset was created with the goal of minimising biases in the data,
since some VQA systems are suspected to exploit them to find answers instead of actually reasoning about the question and scene information~\citep{johnson2016}.

Each CLEVR image depicts a scene with objects in it.
The objects differ by the values of their attributes, 
which are $size$ (big, small), $color$ (brown, blue, cyan, gray, green, purple, red, yellow), $material$ (metal, rubber), and $shape$ (cube, cylinder, sphere). 
Every image comes with a ground truth scene graph describing the scene depicted in it.
Figure~\ref{fig:three_clevr_scenes} contains three images from the CLEVR validation dataset with corresponding questions.

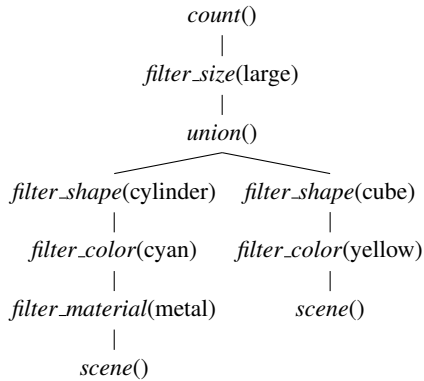
\begin{figure}
\begin{center}
\begin{tikzpicture}[scale=1,level distance=22pt]
\Tree [ .\textit{count}() [ .\textit{filter\_size}(large) [ .\textit{union}() [ 
.\textit{filter\_shape}(cylinder) [ .\textit{filter\_color}(cyan) [ .\textit{filter\_material}(metal) \textit{scene}() ] ] ]
[ .\textit{filter\_shape}(cube)
[ .\textit{filter\_color}(yellow) [  .\textit{scene}() ] ] 
] ] ] ]
\end{tikzpicture}
\end{center}
\caption{CLEVR functional program representing the question: ``How many large things are either cyan metallic cylinders or yellow blocks?''.}\label{fig:fp}
\end{figure}

In CLEVR, questions are constructed using \emph{functional programs} that represent the questions in a structured format. These programs are symbolic templates for a question which are instantiated with the corresponding values. For each such question template, there are one or more natural language sentences to which they are mapped. 
For illustration, the question ``How many large things are either cyan metallic cylinders or yellow blocks?'' from Fig.~\ref{fig:three_clevr_scenes} can be represented by the functional program shown in Fig.~\ref{fig:fp}. There,
function \textit{scene}() returns the set of objects of the scene, the \textit{filter\_*} functions restrict a set of objects to subsets with respective properties, \textit{union}() yields the union of two sets, and \textit{count}() finally returns the number of elements of a set.
A detailed description of functional programs in CLEVR can be found
in the dataset documentation~\citep{johnson2016}.

\section{The VQA Pipeline}{\label{sec:framework}}

\begin{figure}
\centering
\includegraphics[width=\textwidth]{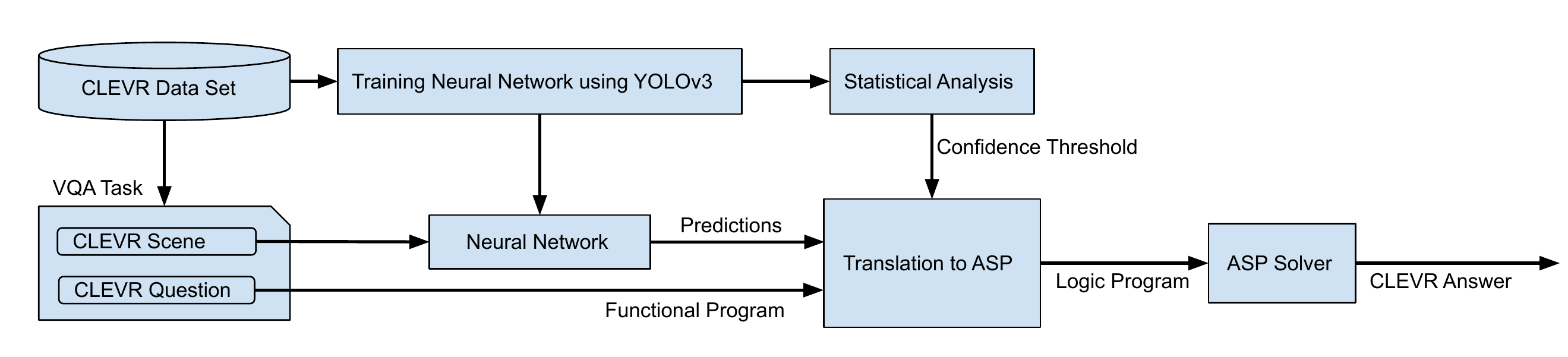}
\caption{Overview of our neuro-symbolic VQA pipeline.}\label{fig:pipeline}
\end{figure}

The architecture of our neuro-symbolic VQA pipeline that builds on object detection and ASP solving is depicted in Fig.~\ref{fig:pipeline}. A particular VQA task, which consists of a CLEVR scene and question, is translated to an ASP program given predictions from a neural network for object detection, a functional program, and a confidence threshold; by running an ASP solver, the answer to the CLEVR task is then figured out.  

Before going into details of our VQA pipeline, we recapitulate the necessary stages of establishing it for the CLEVR dataset:

\begin{enumerate}
    \item {\bf Object detection}: we train neural networks for bounding-box prediction and object classification of the CLEVR scenes; 
    \item {\bf Confidence thresholds}: we determine a threshold for 
    network predictions that we consider to be of high confidence by statistical analysis on the distribution of prediction values of the neural networks;
    \item {\bf ASP encoding}: we translate CLEVR functional programs that represent questions as well as network predictions that pass confidence thresholds into ASP programs and use an ASP solver to compute the answers.
\end{enumerate}

While the VQA tasks are designed to always have a unique answer, the ASP solver may give multiple results that correspond to alternative interpretations of the scene through the object detection network.

\subsection{Object Detection}

We use YOLOv3~\citep{Redmon17} for bounding-box prediction and object classification, 
adopting that the object detector's output is a 
matrix whose rows correspond to the bounding-box predictions in the input picture. 
Each bounding-box prediction is a vector of the form $(c_1,\ldots,c_n,x_1,y_1,x_2,y_2)$, where 
the pairs $(x_1,y_1)$ and $(x_2,y_2)$ give the top-left and bottom-right corner point of the bounding box, respectively, 
and $c_1,\ldots,c_n$ are \emph{class confidence scores} with $c_i \in [0,1]$ for $1 \leq i \leq n$; as customary, higher confidence scores represent higher confidence of a correct prediction. Each $c_i$ represents the score for a specific combination of object attributes $size$, $color$, $material$ and $shape$ and their respective values; we call this combination the \emph{object class} of position $i$. 
For any object class $c$, let 
$\bar{c}$ be the list $\mathit{size}, {shape}, {material}, {color}$ of its attribute values.
For example, assume $c$ is the object class ``large red metallic cylinder'', then 
$\bar{c} = \mathrm{large}, \mathrm{cylinder}, \mathrm{metallic}, \mathrm{red}$.
In total,
there are $n=96$ object classes in CLEVR.

Every row of the prediction matrix has also its own bounding-box confidence score.
The number of bounding-box predictions of the object detection system
depends on the \emph{bounding-box threshold}, which is a hyper-parameter used to 
filter out rows with a low confidence score. 
For example, setting this threshold to $0.5$ discards all predictions with confidence score below $0.5$. 

\subsection{Confidence Thresholds}\label{sec:confthres}

Given class confidence scores $c_1,\ldots,c_n$ from an object detection prediction, we would like to focus on classifications that have reasonable high confidence and discard others with low confidence for the subsequent reasoning process. Using a fixed threshold hardly achieves this, since it does not take the distribution of confidence scores in the application area (or validation data for experiments) into account. 
Our approach solves this problem by fixing the  threshold based on the mean and the standard deviation of prediction scores.

More formally, given a list of prediction matrices $\bm{X}^1,\ldots,\bm{X}^m$, 
where any $\bm{X^i}$ is of dimension $N^i \times M$, {$N^i$ is the number of bounding box predictions in the input image $i$, each of which is described by $M$ features.}
We compute the mean $\mu$  and standard deviation $\sigma$ for the maximum class confidence scores:
\begin{align}
\mu&=\frac{1}{\sum_{k=1}^{m}N^k}\sum\limits_{k=1}^{m}\sum\limits_{i=1}^{N^k}\underset{1 \leq j \leq n}{\max}(\bm{X}^{k}_{i,j}) \label{eq:conf_mean} \\
\sigma&=\sqrt{\frac{1}{\sum_{k=1}^{m}N^k}\sum_{k=1}^{m}\sum_{i=1}^{N^k}(\underset{1 \leq j \leq n}{\max}(\bm{X}^{k}_{i,j})-\mu)^2} \label{eq:conf_sd} 
\end{align}
We suggest computing these values on the validation dataset used in training the object detector. 
Then, we define the \emph{confidence threshold} $\theta$ that determines what is considered a confident class prediction as follows:
\begin{align}
\theta&=\mathit{max}(\mu - \alpha\cdot\sigma, 0)\label{eq:conf_thres}
\end{align}
We consider class predictions as sufficiently confident if
their confidence score is not lower than the mean minus $\alpha$ many standard deviations.
The value for $\alpha$ in Eqn.~(\ref{eq:conf_thres}) is a parameter we call the {\em confidence rate}. It must be provided
and should depend on how well the network is trained:
for a fixed $\alpha$, the number of class predictions that pass the threshold decreases if the network gets better trained as the standard deviation becomes smaller. 
For a fixed network, the number of class predictions that pass the threshold decreases if $\alpha$ decreases and increases otherwise.

\subsection{ASP Encoding}

To solve VQA tasks, we rely on ASP to infer the right answer given 
the neural network output and a confidence threshold. We outline the details
in the following.

\subsubsection{Question encoding}

The first step of our approach is to translate the functional program representing a natural language question into an ASP \emph{fact representation}. 
We illustrate this for the question ``How many large things are either cyan metallic cylinders or yellow blocks?'' in Section~\ref{sec:clevr}.
The respective functional program 
in Fig.~\ref{fig:fp} is encoded by the following ASP facts:
\begin{center}
\begin{minipage}{0.55\textwidth}
\begin{tabular}{rr}
\multicolumn{2}{c}{
\begin{tabular}{r}
$\mathit{end}(8)\,.$ \\
$\mathit{count}(8,7)\,.$ \\
$\mathit{filter\_large(7,6)}\,.$ \\
$\mathit{union}(6,3,5)\,.$
\end{tabular}}\\
$\mathit{filter\_cylinder}(3,2)\,.$ & $\mathit{filter\_cube}(5,4)\,.$\\
$\mathit{filter\_cyan}(2,1)\,.$  & $\mathit{filter\_yellow}(4,0)\,.$\\
$\mathit{filter\_metal}(1,0)\,.$& \\
\multicolumn{2}{c}{\qquad$\mathit{scene}(0)\,.$} 
\end{tabular}
\end{minipage}
\end{center}
The structure of the functional program
is encoded using indices that refer to output (first arguments) and input (remaining arguments) of the respective basic functions.

\subsubsection{Scene encoding}

Let $\bm{X}$ be a prediction matrix and $\theta$ be a confidence threshold 
as described in Section~\ref{sec:confthres}.
Recall that the output of the basic CLEVR function $\mathit{scene}()$ corresponds
to the objects detected in the scene which in turn correspond to the individual rows of $\bm{X}$.

For a row $\bm{X}_i$ with confidence class scores $c_1,...,c_n$,
set $C_i$ contains
every object class $c_j$ with score greater or equal than $\theta$.
If no such $c_j$ exists, then $C_i$ contains
the $k$
classes with highest class confidence scores, where  
$k \in \{1,\ldots,96\}$, is a fixed integer;
intuitively, 
$k$ is a fall-back parameter
to ensure that some classes are selected if all scores are low.

For every row $\bm{X}_i$ with bounding-box corners $(x_1,y_1)$ and $(x_2,y_2)$,
as well as $C_i = \{c_1, \ldots c_l\}$,
we construct a choice rule of form
\begin{align*}
\{ \mathit{obj}(O,i,\bar{c_1},x_1,y_1,x_2,y_2) ; 
   \ldots
   ; \mathit{obj}(O,i,\bar{c_l},x_1,y_1,x_2,y_2) \}=1 \asplarrow \mathit{scene}(O).
\end{align*}

Every object with sufficiently high confidence score will thus be considered for computing the final answer in a non-deterministic way.
For every $c \in C_i$, we add the weak constraint 
\begin{align*}
&\weaklarrow \mathit{obj}(O,i,\bar{c},x_1,y_1,x_2,y_2)\ [w_{c},i]
\end{align*}
where the weight $w_{c}$
is defined as $\min(-1000 \cdot \ln(s), 5000)$, and $s$ is 
the class confidence score for $c$ in $\bm{X}_i$. 
This approach, which comes from the NeurASP implementation, 
achieves that object selections are penalised by a weight which corresponds to the object's class confidence score. Resulting answer sets can thus be ordered according to the total confidence of the involved object predictions.
We refer to this encoding as \emph{non-deterministic scene encoding}, but we also consider 
the 
special case of a \emph{deterministic scene encoding} where each $C_i$
holds only the single object class with the highest confidence score.

\subsubsection{Encoding of the basic CLEVR functions}

We next present encodings for the remaining CLEVR functions.

\smallskip
\noindent
\emph{Filter rules:}\quad
The CLEVR filter rules restrict sets of objects. We only present the rule for
$\mathit{filter\_color}(\mathrm{yellow})$; the rules for the other colours, materials, and shapes are analogous:

\vspace{-0.5\baselineskip}

{
\begin{align*}
    \mathit{obj}(T,I,\ldots,\mathrm{yellow},\ldots) \asplarrow\  \mathit{filter\_yellow}(T,T_1), \mathit{obj}(T_1,I,\ldots,\mathrm{yellow},\ldots)\,. 
\end{align*}
}

The variables $T$ and $T_1$ are used to indicate output resp.\ input references; $I$ represents the object identifier. We omit arguments irrelevant for the particular filter functions.

\smallskip
\noindent
\emph{Count rule:}\quad 
Function $\mathit{count()}$ returns the number of elements of a given set.
We encode it as follows:

\vspace{-0.5\baselineskip}

{
\begin{align*}
\mathit{int}(T,V) \asplarrow \mathit{count}(T,T_1),\ \#\mathit{count}\{I:obj(T_1,I,...)\}=V.
\end{align*}
}
Here, $\#\mathit{count}$ is an ASP \emph{aggregate function} that computes the numbers of object identifiers referenced by variable $T_1$.

\smallskip
\noindent
\emph{Rules for set operations:}\quad 
The two set operation functions in CLEVR are intersection and union. We present respective ASP rules for each of them:

\vspace{-0.5\baselineskip}

{
\begin{align*}
\mathit{obj}(T,I,\ldots) &\asplarrow \mathit{and}(T,T_1,T_2),\ \mathit{obj}(T_1,I,\ldots),\ \mathit{obj}(T_2,I,\ldots)\,. \\
\mathit{obj}(T,I,\ldots) &\asplarrow \mathit{or}(T,T_1,T_2),\ \mathit{obj}(T_1,I,\ldots)\,. \\
\mathit{obj}(T,I,\ldots) &\asplarrow \mathit{or}(T,T_1,T_2),\ \mathit{obj}(T_2,I,\ldots)\,. \\
\end{align*}
}

\vspace{-0.5\baselineskip}

\noindent
\emph{Uniqueness constraint:}\quad 
The CLEVR function $\mathit{unique}()$ is used to assert that there is exactly one input object, 
which is then propagated to the output.
We encode this in ASP using a constraint to eliminate answer sets
violating uniqueness and a rule for propagation:

\vspace{-0.5\baselineskip}

{
\begin{align*}
    &\asplarrow \mathit{unique}(T,T_1),\ \mathit{obj}(T_1,I,\ldots),\ \mathit{obj}(T_1,I',\ldots),\ I \not = I'. \\
    &\mathit{obj}(T,\ldots) \asplarrow \mathit{unique}(T,T_1),\ \mathit{obj}(T_1,\ldots)\,.
\end{align*}
}

\noindent
\emph{Spatial-relation rules:}\quad 
Several CLEVR functions allow to determine objects 
in a certain spatial relation with another object. 
We present the rule 
for identifying all objects that are left relative to a given reference; the rules for right, front, and behind, are 
analogous:

\vspace{-0.5\baselineskip}

{
\begin{align*}
    \mathit{obj}(T,I,\ldots) \asplarrow&\   \mathit{relate\_left}(T,T_1,T_2), I \not = I', X_1<X_1',\,\\
    &\mathit{obj}(T_1,I,\ldots,X_1,\ldots), \mathit{obj}(T_2,I',\ldots,X_1',\ldots)\,.
\end{align*}
}

\noindent
\emph{Exist rule:}\quad 
The $\mathit{exist}()$ rule in CLEVR returns true if the referenced set of objects is not empty,
{
and it returns false otherwise using default negation.
}
Respective ASP rules look as follows:

\vspace{-0.5\baselineskip}

{
\begin{align*}
\mathit{bool}(T,\mathrm{true}) \asplarrow\ & \mathit{exist}(T,T_1),\ \mathit{obj}(T_1,\ldots)\,. \\
\mathit{bool}(T,\mathrm{false}) \asplarrow\ & \mathit{exist}(T,T_1),\ \mathit{not}\ \mathit{bool}(T,\mathrm{true})\,.
\end{align*}
}

\noindent
\emph{Query rules:}\quad 
Query functions allow to return an attribute value of a referenced object. We present a rule to query for the size of an object; the rules for colour, material, and shape look similar:

\vspace{-\baselineskip}

{
\begin{align*}
\mathit{size}(T,\mathit{Size}) \asplarrow\ & \mathit{query\_size}(T,T_1),
 \mathit{obj}(T_1,\ldots,\mathit{Size},\ldots)\,. 
\end{align*}
}

\noindent
\emph{Same-attribute-relation rules:}\quad 
Similar to the spatial-relation functions, same-attribute-relation rules
allow to select sets of objects if they agree on a specified attribute with a specified reference object. We illustrate the ASP encoding for the size attribute, the ones for colour, material and shape are defined with the necessary changes:

\vspace{-0.5\baselineskip}

{
\begin{align*}
\mathit{obj}(T,I,\ldots) \asplarrow&\  \mathit{same\_size}(T,T_1,T_2),\ \mathit{obj}(T_1,I,\ldots, \mathit{Size}, \ldots),\\ 
                                   &\  \mathit{obj}(T_2,I',\ldots, \mathit{Size},\ldots), I \not = I'\,. 
\end{align*}
}

\noindent
\emph{Integer-comparison rules:}\quad 
CLEVR supports the common relations for comparing integers like 
``equals'', ``less-than'', and ``greater-than''. 
We present the ASP encoding for ``equals'':

\vspace{-0.5\baselineskip}

{
\begin{align*}
\mathit{bool}(T,\mathrm{true}) \asplarrow\ & \mathit{equal\_integer}(T,T_1,T_2),\ \mathit{int}(T_1,V), \ \mathit{int}(T_2,V)\,. \\
\mathit{bool}(T,\mathrm{false}) \asplarrow\ & \mathit{equal\_integer}(T,T_1,T_2),\ \mathit{not}\ \mathit{bool}(T,\mathrm{true})\,. \\
\end{align*}
}

\noindent
\emph{Attribute-comparison rules:}\quad 
To check whether two objects have the same attributes, like size, colour, material, or shape,
CLEVR provides attribute-comparisons rules.
The one for size can be represented in ASP as follows, the others are defined analogously:

\vspace{-0.5\baselineskip}

{
\begin{align*}
\mathit{bool}(T,\mathrm{true}) \asplarrow\ & \mathit{equal\_size}(T,T_1,T_2),\ \mathit{size}(T_1,V),\ \mathit{size}(T_2,V)\,. \\
\mathit{bool}(T,\mathrm{false}) \asplarrow\ & \mathit{equal\_size}(T,T_1,T_2),\ \mathit{not}\ \mathit{bool}(V,\mathrm{true})\,. 
\end{align*}
}

In addition to the rules above, we also use rules to derive the $ans/1$ atom that extracts the final answer for the encoded CLEVR question from the output of the basic function at the root of the computation:

{
\begin{align*}
&\mathit{ans}(V) \asplarrow \mathit{end}(T), \mathit{size}(T,V)\,. && \mathit{ans}(V) \asplarrow \mathit{end}(T), \mathit{shape}(T,V)\,. \\
&\mathit{ans}(V) \asplarrow \mathit{end}(T), \mathit{color}(T,V)\,. && \mathit{ans}(V) \asplarrow \mathit{end}(T), \mathit{bool}(T,V)\,. \\
&\mathit{ans}(V) \asplarrow \mathit{end}(T), \mathit{material}(T,V)\,. && \mathit{ans}(V) \asplarrow \mathit{end}(T), \mathit{int}(T,V)\,.\\
&\asplarrow \mathit{not}\ \mathit{ans}(\_).
\end{align*}
}

\noindent
The last constraint enforces that at least one answer is derived.

Putting all together, to find an answer to a CLEVR question, we translate the corresponding functional program into its fact representation and join it with the rules from above.
Each answer set 
then 
corresponds to a CLEVR answer
founded in a particular choice for scene objects.
For the deterministic, resp., non-deterministic encoding, at most one, resp., multiple answer sets are possible; no answer set means imperfect object recognition.
In case of multiple answer sets, we use answer-set optimisation over the weak constraints to 
determine the most plausible solution.

\begin{table}
\centering 
\caption{Precision and recall  for object detection on CLEVR scenes.}\label{tab:results_scene_encoding}
{\tablefont\begin{tabular}{@{\extracolsep{\fill}}lrrr@{\qquad}rrr}
\topline
\textbf{Epochs/Threshold} &
 25/0.25 & 50/0.25 & 200/0.25 & 25/0.50 & 50/0.50 & 200/0.50  
\midline
recall           & $71.23\%$ & $96.39\%$ & $99.20\%$ & $44.41\%$ & $89.06\%$ & $98.03\%$ \\
precision        & $83.77\%$ & $98.27\%$ & $99.58\%$ & $97.17\%$ & $99.28\%$ & $99.82\%$ 
\botline
\end{tabular}}
\end{table}

\section{Experiments on the CLEVR Dataset}{\label{sec:experiments}}

Recall that the parameters of our approach are (i) the 
bounding-box threshold for object detection, (ii) 
the confidence rate $\alpha$ for computing the confidence threshold as distance from the mean in terms of standard deviations,
and (iii) $k$ as a fall-back parameter for object-class selection.
We experimentally evaluated our approach on the CLEVR dataset to study the effects of different
parameter settings. In particular, we study
\begin{itemize}
\item different bounding-box thresholds and training epochs for object detection,
\item how the deterministic scene encoding compares to the non-deterministic one, 
and
\item runtime performance of our approach in comparison with NeurASP and 
ProbLog.
\end{itemize}
For the  non-deterministic scene encoding, we consider different settings for $\alpha$  and set $k=1$.

{
We restricted our experiments to a sample of 15000 CLEVR questions as the systems NeurASP and ProbLog would exceed the memory limits on the unrestricted dataset.  
}

All experiments were carried out on an Ubuntu (20.04.3 LTS) system with a 3.60GHz Intel CPU, 16GiB of RAM, and an NVIDIA GeForce GTX 1080 GPU with 8GB of memory installed.

\subsection{Object-Detection Evaluation}

For object detection, we used  an open-source implementation of YOLOv3~\citep{Redmon17}.\footnote{\url{https://github.com/eriklindernoren/PyTorch-YOLOv3}.} 
The system was trained on $4000$ CLEVR images with bounding box annotations, as suggested in 
related work~\citep{Yi18}.
We used models trained for $25$, $50$ and $200$ epochs, resp., to obtain different levels of training for the neural networks.
For the  bounding-box thresholds, we considered two settings, namely $0.25$ and $0.50$.
Table~\ref{tab:results_scene_encoding} summarises the 
results of the evaluation of how the networks of different training quality perform for detecting the objects in the CLEVR scenes. We report on   
precision and recall, which are defined as usual in terms of 
true positives (TP), false positives (FP), and false negatives (FN).
A {\em TP}\/  (resp., {\em FP}) is a prediction 
that is correct (resp., incorrect) w.r.t.\ the scene annotations in CLEVR.
An \emph{FN} is an object  that exists according to the scene annotations,
but there is no corresponding prediction. 

As expected, our results show that the total number of FP and FN decreases for the better trained YOLOv3 models. Naturally, a low bounding-box threshold yields more FP detections, while the number of FN decreases. Setting the bounding-box threshold to a higher value usually leads to fewer FP but also more FN. 

\subsection{Question-Answering Evaluation}

\begin{table}
\centering
\caption{Results for CLEVR question answering.}\label{tab:results_question_answering}
{\tablefont\begin{tabular}{@{\extracolsep{\fill}}rlrrr@{\quad}rrr}
\topline
\multicolumn{3}{l}{\textbf{Epochs/Threshold}} 
 25/0.25 & 50/0.25 & 200/0.25 & 25/0.50 & 50/0.50 & 200/0.50 
\midline
\multicolumn{8}{l}{\textbf{Deterministic Scene Encoding}} \\[.5em]
\multirow{3}{*}{ } &
correct &  $64.39\%$ &  $92.93\%$ &  $97.01\%$ &  $42.33\%$ &  $82.79\%$ & $95.05\%$ \\
& wrong   & $17.60\%$ & $3.06\%$  & $1.11\%$  & $29.55\%$ & $8.65\%$  & $2.22\%$ \\
& no answer & $18.01\%$ & $4.01\%$ & $1.88\%$  & $28.12\%$ & $8.56\%$  & $2.73\%$ \\
\hline
\multicolumn{8}{l}{\textbf{Non-Deterministic Scene Encoding}} \\[.5em]
\multirow{3}{*}{$\alpha=0.5$} &
correct & $74.12\%$ & $93.13\%$ & $96.94\%$ & $72.36\%$ & $92.54\%$ & $96.71\%$ \\
& wrong   & $12.29\%$ & $2.60\%$  & $1.01\%$  & $13.21\%$ & $2.98\%$ & $1.27\%$ \\
& no answer & $13.59\%$ & $4.27\%$  & $2.05\%$  & $14.43\%$ & $4.48\%$ & $2.02\%$ \\
\hline
\multirow{3}{*}{$\alpha=1.0$} &
correct & $76.17\%$ & $93.13\%$ & $96.94\%$ & $74.29\%$ & $92.54\%$ & $96.71\%$ \\
& wrong   & $12.53\%$ & $2.60\%$  & $1.01\%$  & $13.47\%$ & $2.98\%$ & $1.27\%$ \\
& no answer & $11.30\%$ & $4.27\%$  & $2.05\%$  & $12.24\%$ & $4.48\%$ & $2.02\%$ \\
\hline
\multirow{3}{*}{$\alpha=1.5$} &
correct & $80.61\%$ & $93.13\%$ & $96.94\%$ & $78.83\%$ & $92.55\%$ & $96.71\%$ \\
& wrong   & $13.02\%$ & $2.60\%$  & $1.01\%$  & $14.00\%$ & $2.98\%$ & $1.27\%$ \\
& no answer & $6.37\%$ & $4.27\%$  & $2.05\%$  & $7.17\%$ & $4.47\%$ & $2.02\%$ \\
\hline
\multirow{3}{*}{$\alpha=2.0$} &
correct & $84.09\%$ & $93.17\%$ & $96.94\%$ & $82.65\%$ & $92.56\%$ & $96.71\%$ \\
& wrong   & $13.52\%$ & $2.60\%$  & $1.01\%$  & $14.53\%$ & $2.98\%$ & $1.27\%$ \\
& no answer & $2.39\%$ & $4.23\%$  & $2.05\%$  & $2.82\%$ & $4.46\%$ & $2.02\%$
\botline
\end{tabular}}
\end{table}

We used the ASP solver \texttt{clingo} (v.~5.5.1) to compute answer sets.%
\footnote{\url{https://potassco.org/clingo/}.}
Table~\ref{tab:results_question_answering} sheds light on the 
impact of the training level and bounding-box thresholds of the models on question answering for the deterministic and non-deterministic scene encodings.
Our system yields either correct, incorrect, or no answers to the CLEVR questions, and we report respective rates.
The non-deterministic scene encoding outperformed the deterministic approach for all settings of training epochs and bounding-box thresholds, and the rate of correct answers increases with larger $\alpha$.
The differences are considerable if the networks are trained rather poorly and 
and become small or even disappear for well trained ones.
It thus seems beneficial to consider more than one prediction of the object detection system 
if network predictions are less than perfect.
Also, lower bounding-box thresholds lead to more correct results in all
cases, especially for the deterministic encoding. Hence, being too
selective in the object detection is counterproductive.

\subsection{Comparison with NeurASP and ProbLog}

\begin{table}
\centering
\caption{Comparisons with NeurASP and ProbLog. 
}\label{tab:results_NeurASP}
{\tablefont\begin{tabular}{@{\extracolsep{\fill}}lrrr@{\quad}rrr}
\topline
\textbf{Epochs/Threshold} &
 25/0.25 & 50/0.25 & 200/0.25 & 25/0.50 & 50/0.50 & 200/0.50  
\midline
\multicolumn{7}{l}{\textbf{NeurASP}} \\
correct & $86.09\%$ & $96.74\%$ & $98.53\%$ & $84.87\%$ & $96.17\%$ & $98.20\%$ \\
wrong & $13.88\%$ & $3.21\%$ & $1.45\%$ & $15.09\%$ & $3.77\%$ & $1.77\%$ \\
no answer & $0.03\%$ & $0.05\%$ & $0.03\%$ & $0.04\%$ & $0.06\%$ & $0.03\%$ \\
\hline
\multicolumn{7}{l}{\textbf{NeurASP \textup{(}best prediction\textup{)}}} \\
correct & $75.63\%$ & $95.78\%$ & $98.33\%$ & $53.12\%$ & $88.21\%$ & $96.87\%$ \\
wrong & $23.51\%$ & $4.14\%$ & $1.63\%$ & $38.17\%$ & $11.30\%$ & $3.06\%$ \\
no answer & $0.87\%$ & $0.08\%$ & $0.03\%$ & $8.71\%$ & $0.49\%$ & $0.07\%$ \\
\hline
\multicolumn{7}{l}{\textbf{ProbLog}} \\
correct & $83.84\%$ & $96.25\%$ & $98.33\%$ & $81.97\%$ & $95.27\%$ & $97.85\%$ \\
wrong & $14.46\%$ & $2.66\%$ & $1.14\%$ & $15.45\%$ & $3.13\%$ & $1.34\%$ \\
no answer & $1.70\%$ & $1.09\%$ & $0.53\%$ & $2.57\%$ & $1.59\%$ & $0.81\%$ 
\botline
\end{tabular}}
\end{table}

\begin{table}
\centering
\caption{Comparison of total VQA runtimes (in seconds). 
}\label{tab:results_runtimes}
{\tablefont\begin{tabular}{@{\extracolsep{\fill}}ll@{}rrr@{\quad}rrr}
\topline
\multicolumn{2}{l}{\textbf{Epochs/Threshold}} &
 25/0.25 & 50/0.25 & 200/0.25 & 25/0.50 & 50/0.50 & 200/0.50  
\midline
\multicolumn{2}{l@{}}{\textbf{NeurASP}} & $9149$ & $4375$ & $4364$ & $8750$ & $4234$ & $4694$ \\
\multicolumn{2}{l@{}}{\textbf{NeurASP (best prediction)}} & $3114$ & $3628$ & $3575$ & $1245$ & $3174$ & $3577$ \\ 
\multicolumn{2}{l@{}}{\textbf{ProbLog}} & $6235$ & $6894$ & $6463$ & $5311$ & $5883$ & $6138$ \\
\multicolumn{2}{l@{}}{\textbf{Deterministic Scene Encoding}} & $84$ & $89$ & $87$ & $81$ & $89$ & $85$ \\
\multirow{4}{*}{\textbf{Non-Det.~Scene Encoding  with}} & $\alpha=0.5$ & $112$ & $123$ & $130$ & $109$ & $124$ & $131$ \\
& \textbf{$\alpha=1.5$} & $141$ & $125$ & $126$ & $140$ & $125$ & $133$ \\
& \textbf{$\alpha=2.0$} & $154$ & $126$ & $130$ & $165$ & $125$ & $135$ \\
& \textbf{$\alpha=2.5$} & $3656$ & $127$ & $128$ & $3541$ & $123$ & $128$ 
\botline
\end{tabular}}
\end{table}

The related systems NeurASP and DeepProbLog
also embody the idea of non-determinism for object classifications, but
they do not incorporate a mechanism to restrict object classes to ones with high confidence like in our approach; this can drag down performance {considerably}. We conducted different experiments to investigate the impact of limiting the number of object classes as in our approach on runtimes and accuracy of the questions answering task in comparison with the aforementioned related systems.

The choice rules in NeurASP always contain the 96 CLEVR object classes,
whose scores come from the YOLOv3 network.
In addition, we also used a setup for NeurASP where only the highest prediction is considered while the probabilities of all other atoms are set to~0. 
While the former setting is more similar to our approach, the latter is the one 
used by \cite{yang2020neurasp} in their object detection example.
Table~\ref{tab:results_NeurASP} summarises the results on question answering accuracy, and
Table~\ref{tab:results_runtimes} shows the total runtime for the different systems under consideration on the 15000 questions. 
NeurASP outperforms our approach in terms of correct answers as it does not restrict the number of atoms for the choice rules. However, this comes at a price as 
runtimes are much longer, which can be explained by the inflation of the search space due to the unrestricted choice rules. Also, the rate of incorrect answers is higher for NeurASP while our approach will more often remain agnostic when in doubt.
For the non-deterministic encoding of our approach, we observe a similar jump in runtimes for $\alpha=2.5$ as more object classes are included in the choice rules. 

We could not use DeepProbLog for CLEVR directly as 
annotated disjunctions that depend on a variable number of objects in a scene are not supported
and would require some extensions. 
Instead, we evaluated a translation to ProbLog, as DeepProblog's inference component is essentially the one of 
Problog \citep{manhaeve2018deepproblog}.
Recall that for NeurASP and DeepProbLog, neural network outputs are interpreted as probability distributions. While NeurASP does not strictly require this and works also in our setting, ProbLog is less lenient and network outputs need to be normalised so that the sum of their scores does not exceed~1. This does however not change results as only the relative order of the object-class scores is relevant for determining the most plausible answer. 
For the case of the unrestricted 96 object classes,
computing results on our hardware was infeasible.
We thus considered only the three object classes with highest confidence scores for
every bounding-box prediction in
our experiments.
Results on runtimes and accuracy are therefore lower bounds for the unrestricted case.
The picture for ProbLog is quite similar to that of NeurASP: While additional predictions help for question answering to some extent, this is at the cost of a considerable increase in runtime.

Overall, the experiments further support our belief that non-determinism 
is useful for neuro-symbolic VQA systems and 
suitable mechanisms to restrict it do reasonable choices allows for more efficient implementations.

\section{Further Related Work}{\label{sec:rel}}
Purely deep-learning-based approaches~\citep{yang2016stacked,lu2016hierarchical,jabri2016revisiting} led to significant advances in VQA. Some systems rely on attention mechanisms to focus on certain features of the image and question to retrieve the answer~\citep{yang2016stacked,lu2016hierarchical}. \cite{jabri2016revisiting}) achieved good results by framing a VQA task as a classification problem. 
Some VQA systems are however suspected to not learn to reason but to exploit dataset biases to find answers, as described by \cite{johnson2016}.

Besides these purely data driven attempts, there are also systems which incorporate symbolic reasoning in combination with neural-based methods for VQA~\citep{Yi18,basu2020aqua,mao2019neuro}. 
The system proposed by \cite{Yi18} consists of a scene parser, which retrieves object level scene information from images, a question parser, which creates symbolic programs from natural language questions, and  a program executor that runs these programs on the abstract scene representation. 
Our system is akin to this system, but we use ASP for scene representation as well as question encoding, and our program executor is an ASP solver. 
A similar system architecture appears in 
the approach by \cite{mao2019neuro}
with the difference that scene and question parsing is jointly learned from image and question-answer pairs, whereas the components of \citeauthor{Yi18}'s system are trained separately. This means that annotated images are not necessary for training, which makes the system more versatile. The approach of \cite{basu2020aqua} builds like ours on ASP. They 
use object-level scene representations and parse natural language questions to obtain rules which represent the question. The answer to a question is given by the answer set for the image-question encoding which is combined with commonsense knowledge. However, their approach 
is not amenable to non-determinism for the scene encoding in order to deal with competing object classifications as we do.

{ 
\cite{riley2019integrating} present an integrated approach for deep learning and ASP for  representing incomplete commonsense knowledge and non-monotonic reasoning that also involves learning and program induction. They apply their approach to VQA tasks with explanatory questions and are able to achieve better accuracy on small data sets than end-to-end architectures.
As in our work, they use neural networks to extract features of an image. 
We focus however more narrowly on the interface between the neural network outputs and the logical rules and turn network outputs into choice rules to further improve robustness, which has not been considered by \citeauthor{riley2019integrating}.   
}

\section{Conclusion}{\label{sec:concl}}

We have introduced a neuro-symbolic VQA pipeline for the CLEVR dataset based on
a translation from CLEVR questions in the form of functional programs to ASP rules, where we  proposed a non-deterministic and a deterministic approach to encode object-level scene information. Notably, non-determinism is restricted to network predictions that pass a confidence threshold determined by statistical analysis. It takes the variance of predication quality into account and can be adjusted by the novel confidence rate parameter $\alpha$, which supports control of non-determinism, resp.\, disjunctive information.

Our experiments confirm that, on the one hand, non-determinism is important for robustness of the reasoning component especially if the neural networks for object classification are not well-trained or predictions are negatively affected by other causes. 
On the other hand, unrestricted non-determinism as featured by related neuro-symbolic systems can pose a performance bottleneck. Our method of using a confidence threshold is a viable compromise between quality of question answering and efficiency of the reasoning component. 
{
The insight that it makes sense to deal with uncertainty at the level
of the reasoning component is in fact not restricted to ASP and therefore also provides directions to further improve related approaches.
}

While CLEVR is well-suited for the purposes of this work, the scenes are quite simple and do not require advanced features of ASP like expressing common-sense knowledge. For future work, we intend to apply our approach
also to other datasets, especially ones that do not use synthetic scenes and where object classification might be harder, resulting in increased uncertainty.
{
There, using additional domain knowledge and representing it with ASP could come in handy.}
Furthermore, we plan to extend our pipeline to closed-loop reasoning, i.e., using the output of the ASP solver also in the learning process.



\bibliographystyle{tlplike}
\bibliography{ref}

\end{document}